\documentclass[10pt,twocolumn,letterpaper]{article}

\usepackage{cvpr}              
\definecolor{cvprblue}{rgb}{0.21,0.49,0.74}
\usepackage[pagebackref,breaklinks,colorlinks,allcolors=cvprblue]{hyperref}
\usepackage{booktabs}       
\usepackage{amsfonts}       
\usepackage{nicefrac}       
\usepackage{microtype}      
\usepackage{xcolor}         
\usepackage{adjustbox}
\usepackage{multirow}
\usepackage{arydshln}
\usepackage{subcaption}
\usepackage{wrapfig}
\usepackage{amsmath}
\usepackage{subfiles}
\usepackage{float}

\def\paperID{15860} 
\def\confName{CVPR}
\def\confYear{2026}

\title{VCE: Safe Autoregressive Image Generation
via Visual Contrast Exploitation}

\author{%
    \textbf{Feng Han}$^{1,2}$, \textbf{Chao Gong}$^{1,2}$, \textbf{Zhipeng Wei}$^{3,4}$, \textbf{Jingjing Chen}$^{1,2}$\thanks{Corresponding author.}, \textbf{Yu-Gang Jiang}$^{1,2}$\quad \\
    $^{1}$ Shanghai Key Lab of Intell. Info. Processing, School of CS, Fudan University\\
    $^{2}$ Shanghai Collaborative Innovation Center on Intelligent Visual Computing\\
    $^{3}$ International Computer Science Institute, $^{4}$ UC Berkeley\\
}

\begin{document}

\maketitle




\begin{abstract}
Recently, autoregressive image generation models have demonstrated impressive capabilities in producing highly realistic images. Models such as GPT-4o and LlamaGen can not only produce images that faithfully emulate renowned artistic styles (e.g Van Gogh, Ghibli), but also inadvertently generate Not-Safe-For-Work (NSFW) content, raising significant concerns regarding copyright infringement and ethical misuse. Despite these risks, safety mechanisms for autoregressive text-to-image models remain underexplored. Existing concept erasure methods, are predominantly tailored to diffusion models that operate in denoising latent space or cross-attention layers, which are not directly applicable to autoregressive models that generate images token by token without cross-attention layers. To address this critical gap, we propose \textbf{V}isual \textbf{C}ontrast \textbf{E}xploitation (VCE), a novel framework comprising: (1) a contrastive image-pair construction paradigm that precisely decouples and contrasts unsafe concepts from other semantic content, and (2) a VSafe-DPO training strategy that enhances the model's ability to identify and leverage visual contrasts from image pairs, enabling precise concept erasure. Our comprehensive experiments across three challenging tasks (artist style erasure, object removal and explicit content erasure) demonstrate that our method effectively erases unsafe concepts and maintains the integrity of unrelated safe concepts.
\end{abstract}    
\section{Introduction}
\begin{figure*}[tb] 
    \centering 
    \includegraphics[width=0.9\textwidth]{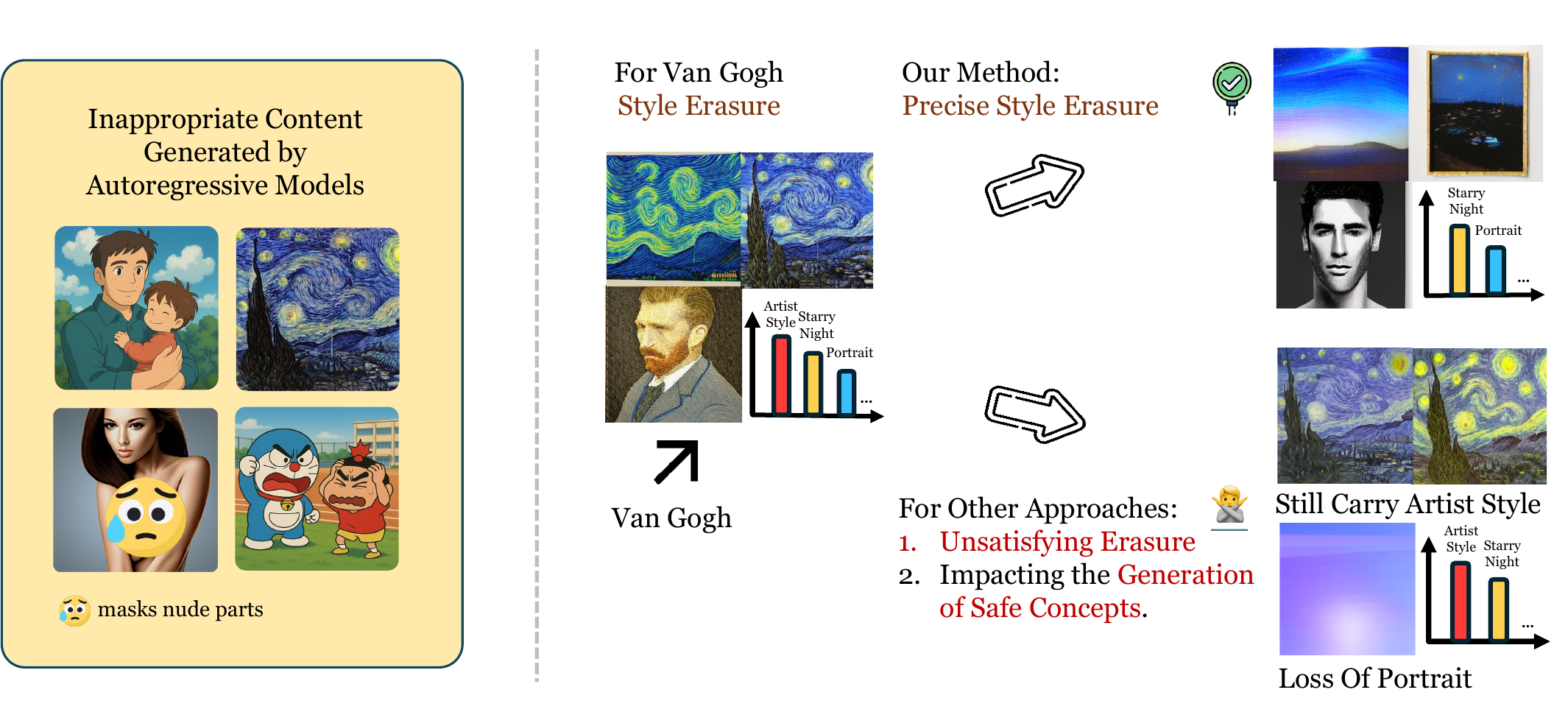} 
    \caption{Left: Autoregressive generative models possess the capability to imitate artistic styles and generate nude and violent images. Right: Our approach precisely decouples and erases the Van Gogh style from generated images.}
    \label{fig:intro} 
    \vspace{-5mm}
\end{figure*}

Recent years have witnessed substantial progress in the field of text-to-image generation, with models exhibiting unparalleled capabilities in generating highly realistic and diverse visual content~\citep{ramesh2021zero, ramesh2022hierarchical}. Diffusion models which employ an iterative denoising process for image generation have dominate the area of text-to-image generation for yeas due to its high fidelity \citep{rombach2022high,esser2024scaling}. Recently, Autoregressive (AR) models which synthesize images via a next-token prediction mechanism, have emerged as the other compelling alternative (e.g. LlamaGen~\citep{sun2024autoregressive} and Janus-Pro~\cite{chen2025janus}), matching or even surpassing diffusion models in generation efficiency and text comprehension \citep{tian2024visual}. Notably, state-of-the-art autoregressive image generation models \citep{sun2024autoregressive, StabilityAI_StableDiffusion2, peebles2023scalable} have exhibited exceptional capabilities in generating images with remarkable fidelity and aesthetic quality. Beyond these advances, these models accurately emulate renowned artistic styles such as Ghibli, Van Gogh, Picasso, and generate images indistinguishable from human-created artwork \citep{team2023gemini,li2024autoregressive}. Meanwhile, they may inadvertently produce inappropriate nude and violent content \citep{schramowski2023safe} (Fig.~\ref{fig:intro}). These phenomena have ignited intense ethical and legal concerns, particularly debates regarding copyright infringement when AI systems replicate specific artistic styles without explicit permission \citep{shan2023glaze,salman2023raising}, highlighting the urgent need for responsible implementation of text-to-image models \citep{schramowski2023safe, rando2022red}.

While both diffusion models and autoregressive models represent dominant paradigms in text-to-image generation, the former has received significantly more attention in recent safety studies.
Early safety approaches, including dataset filtering and model retraining \citep{schuhmann2022laion}, prove inadequate, as even carefully curated and retrained models could still generate unsafe content \citep{qu2023unsafe,schneider2024image}. Subsequently, more advanced methods focus on inference-time interventions that interfere with internal noise latent and cross-attention latent spaces \citep{dong2023towards,schramowski2023safe,yoon2024safree, dong2024towards2, wang2025precise}, as well as parameter fine-tuning approaches that align unsafe concept representations with meaningless text prompts \citep{chavhan2024conceptprune, gandikota2023erasing, gandikota2024unified, gong2024reliable, han2025dumo, li2025speed, zhang2024defensive, gao2025eraseanything, bui2024erasing}. Moreover, recent advancements have further refined these techniques through segmenting unsafe concepts from activation maps \citep{lu2024mace}, constructing contrastive datasets via targeted image editing \citep{park2024direct}, and introducing diverse safe anchor concepts \citep{lyu2024one}. These methods have proven effective at reducing unsafe generations in diffusion models.

Despite these safety achievements in the diffusion setting, safety mechanisms developed for diffusion models face significant limitations when transferred to AR models due to fundamental architectural differences. Firstly, the token-by-token prediction paradigm of AR models \citep{dong2024towards} hinders the transferability of diffusion-specific safety mechanisms that operate on global image latent space. Therefore, the direct transfer of inference-time interventions like SLD \citep{schramowski2023safe} proves ineffective (Fig ~\ref{fig:artistcomparison}). Secondly, most safety mechanisms~\cite{gandikota2024unified, lu2024mace, huang2024receler} achieve concept erasure by manipulating the cross-attention layers of diffusions, which is absent at AR models, making these techniques not compatible. Furthermore, naive implementations of direct fine-tuning, such as maping unsafe text prompt to safe images\citep{gandikota2023erasing}, leads to insufficient erasure of the target concept and impact the generation of safe concepts (Fig.~\ref{fig:intro}). Consequently, there is an urgent need for effective approaches that safeguard AR models without compromising their generative capabilities. 

\begin{figure}[tb]
    \centering
    \begin{adjustbox}{max width=0.8\textwidth}
    \begin{subfigure}[b]{0.48\linewidth}
        \centering
        \includegraphics[width=\textwidth]{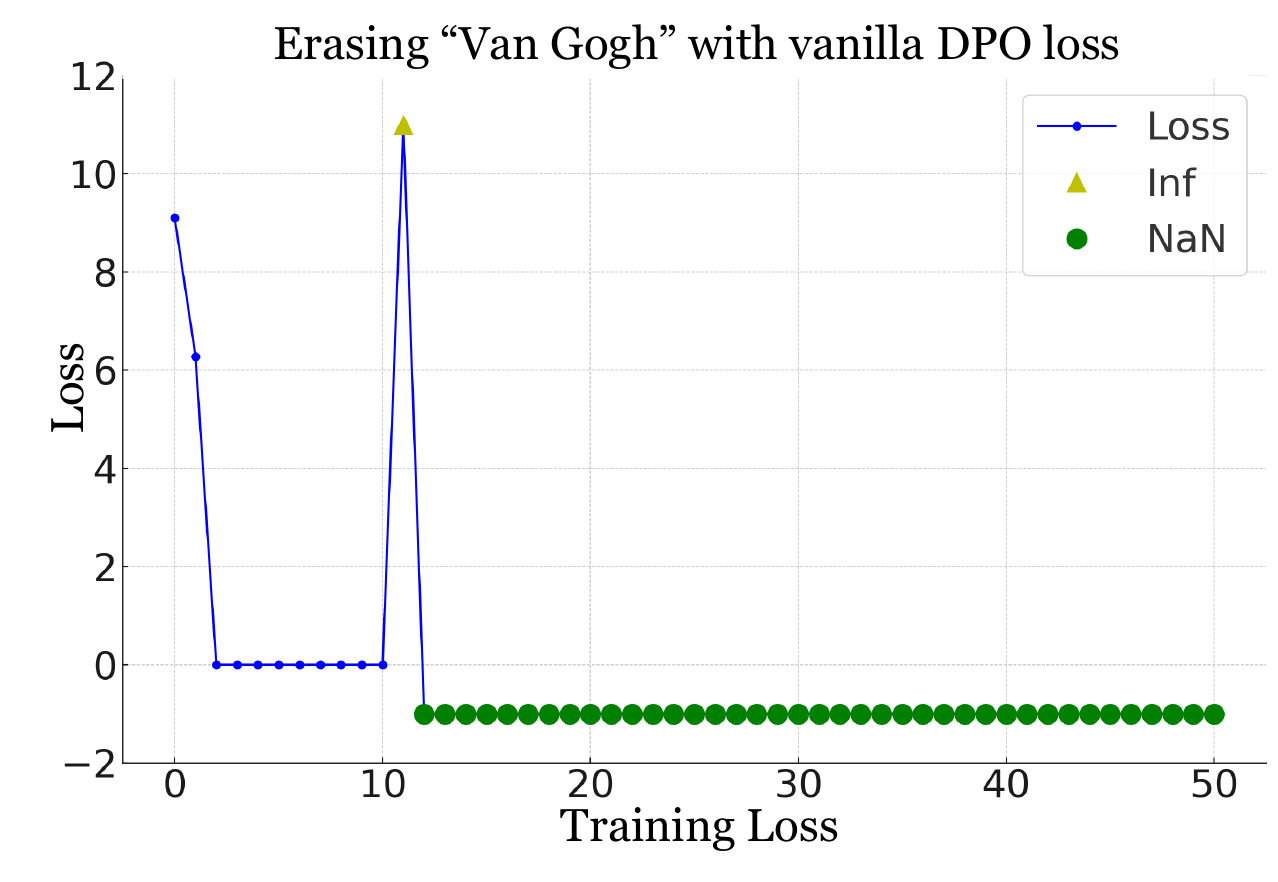} 
    \end{subfigure}
    \hfill
    \begin{subfigure}[b]{0.48\linewidth}
        \centering
        \includegraphics[width=\textwidth]{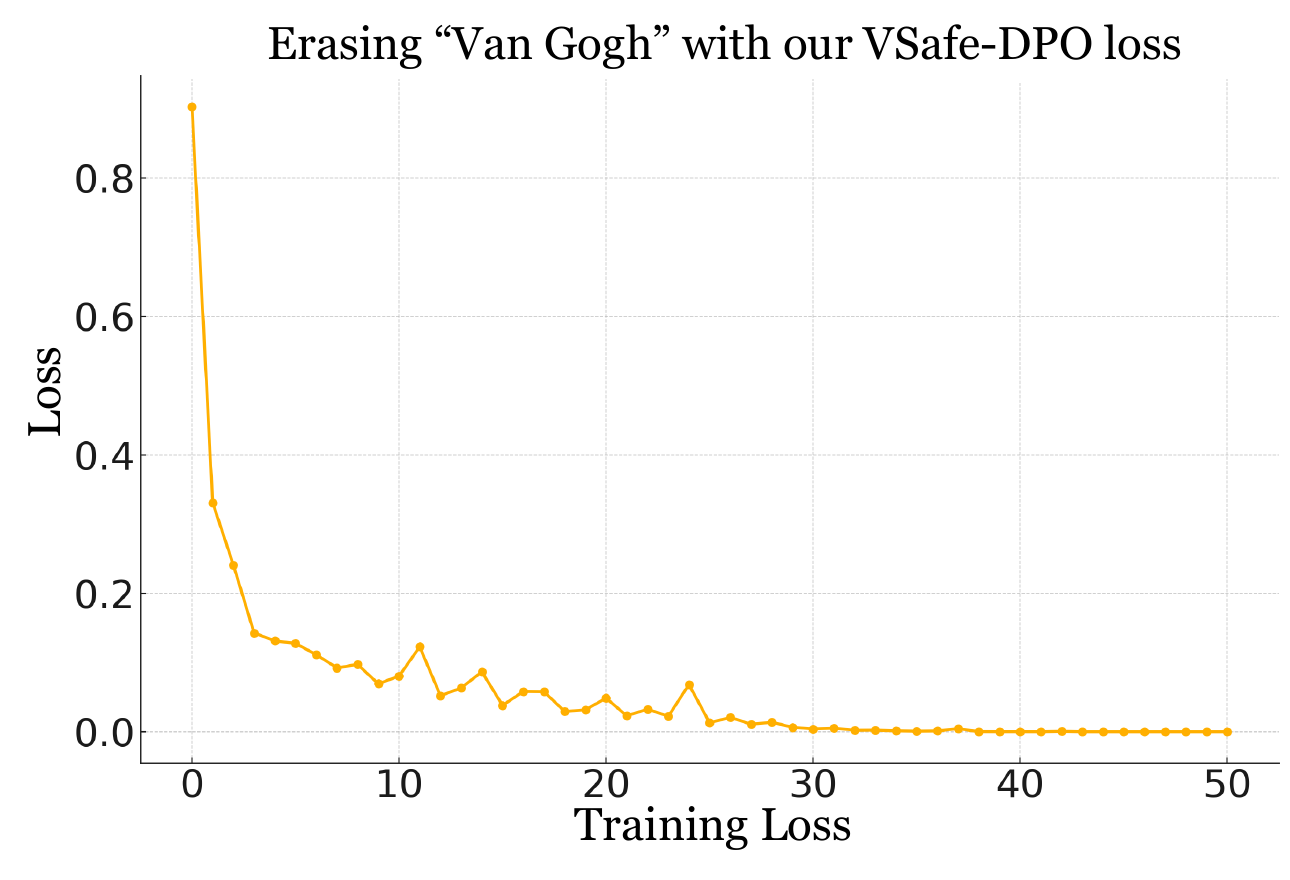} 
    \end{subfigure}
    \end{adjustbox}
    \caption{Training collapse is observed when using the vanilla DPO loss, while our VSafe-DPO loss demonstrates rapid and stable convergence.}
    \label{fig:losscurve}
    \vspace{-4mm}
\end{figure}

To address the limitations of diffusion-specific safety methods and naive fine-tuning method below, we introduce our \textbf{V}isual \textbf{C}ontrast \textbf{E}xploitation (VCE) framework which comprises two key components, the \textbf{contrastive image-pair construction paradigm}, and the \textbf{VSafe-DPO} training methodology.(1) Our data construction begins by leveraging the original AR model~\citep{sun2024autoregressive} to generate images that inherently includes both safe and unsafe visual content. 
We then employ a meticulously designed captioning and filtering approach to selectively disregard only the unsafe concepts, preserving captions describes the semantically safe concepts within these generated images. Subsequently, these refined, safety-aligned captions are utilized to reconstruct the image of non-target, safe content.
Moreover, (2) considering that visual information within these image pairs inherently contrasts unsafe concepts, we employ a token-drop mechanism to diminish textual influence, consequently enhancing the model's reliance on visual cues. 
Additionally, we incorporate token-level average loss to enhance training stability and address optimization challenges in this concept erasure task (Fig.~\ref{fig:losscurve}).

In summary, our contributions are three-fold: (1) To our knowledge, we are the first to explore concept erasure within the next-token-prediction autoregressive image generation paradigm; (2) We propose a novel framework dubbed ``VCE'', which features a contrastive image-pair data construction paradigm and a VSafe-DPO training methodology, including a token-drop mechanism and token-level average loss, to enhance training stability and achieves precise concept erasure; (3) The extensive experiments, including artistic style erasure, object removal and explicit content erasure, validate the effectiveness of our proposed approach in achieving effective concept erasure while maintaining the model’s overall generation capabilities. We get an 82.9\% reduction of generated nude body parts and achieve the best erasure score and $\mathrm{CLIP}_\mathrm{d}$ over all the experiments.

\begin{figure*}[t] 
    \centering 
    \includegraphics[width=0.9\textwidth]{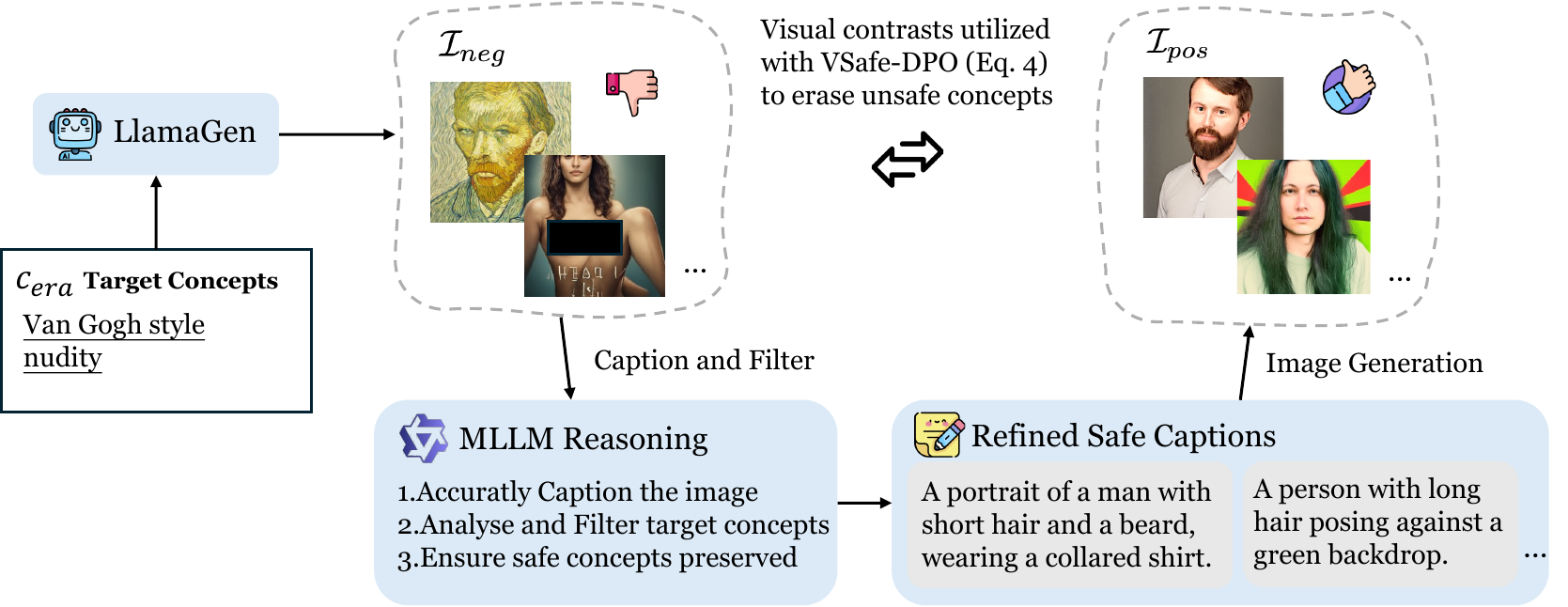} 
    \caption{Framework of our VCE method. We first generate images from target concepts, which model their semantic space. A caption-and-filter process followed by image generation is adopted to generate only safe content, accuratly decoupling the unsafe content. Finally, visual contrasts from these image pairs are exploited through our VSafe-DPO training methodology to erase target concepts.}
    \label{fig:method} 
    \vspace{-5mm}
\end{figure*}
\section{Related Work}

\subsection{Autogressive Generation Models}
Autoregressive (AR) models have recently garnered considerable attention in image generation, particularly following the demonstration of GPT-4o's \citep{hurst2024gpt} remarkable capabilities across diverse domains, from photorealistic imagery to stylized cartoon editing. AR models can be broadly categorized based on whether they operate on discrete or continuous tokens. One approach involves predicting quantized tokens. For example, LlamaGen \citep{sun2024autoregressive} uses the same architecture as VQGAN \citep{esser2021taming} and investigates the scalability of image generation using a pure LLaMA \citep{touvron2023llama} architecture via next-token prediction. In a similar vein, VAR \citep{tian2024visual} introduces next-scale prediction to further enhance generation performance. RandAR \citep{pang2024randar} explores random-order next-token prediction with causal decoder-only models, unlocking zero-shot capabilities such as inpainting. The alternative approach operates on continuous tokens. MAR \citep{li2024autoregressive} employs a lightweight diffusion block to decode continuous latent features. FLUID \citep{fan2024fluid} further demonstrates and scales this approach using a random-order autoregressive scheme on continuous tokens. However, despite their impressive generative abilities, these AR models often struggle to discern copyrightable elements within user-provided prompts, raising potential copyright infringement concerns \citep{CNN_ChatGPT_Ghibli_2025}.

\subsection{Diffusion Safety Mechanisms}
Diffusion safety mechanisms aim to mitigate the generation of inappropriate content by text-to-image (T2I) diffusion models like Stable Diffusion (SD). Initial approaches focus on data filtering \citep{OpenAI_DALLE2_BiasSafety,schuhmann2022laion,StabilityAI_StableDiffusion2}, but retraining on censored datasets is resource-intensive and can potentially introduce biases \citep{OConnor_StableDiffusion1vs2_2022}. Post-generation filtering \citep{NudeNet,rando2022red} and inference-guided methods \citep{schramowski2023safe,yoon2024safree} offer alternative safeguards. However, these methods are susceptible to circumvention \citep{rando2022red}. Recent research has concentrated on fine-tuning techniques to eliminate target concepts from model outputs. Early methods like CA \citep{kumari2023ablating} and ESD \citep{gandikota2023erasing} replace target concepts with designated alternatives. Recognizing the importance of concept preservation, \citep{heng2023selective,gandikota2024unified,bui2025fantastic} propose different optimization to mitigate forgetting.
Subsequent methods \citep{lyu2024one,lu2024mace,han2025dumo,lee2025concept} further introduce additional modules, such as LoRAs or more complex architectures. To enhance the reliability and robustness of erasure, certain methods \citep{gong2024reliable,huang2024receler,zhang2024defensive} incorporate adversarial training. However, these concept erasure methods are designed solely for diffusion models and, as we observe, cannot be readily adapted to ARs. While DUO \citep{park2024direct} employs DPO \citep{wallace2024diffusion} for safe diffusion, it exhibits limitations in style erasure because its reliance on image editing for preference pair construction proves adequate for localized modifications but falls short in achieving comprehensive style removal. Therefore, the development of concept erasure methods explicitly tailored for ARs and capable of addressing a diverse range of inappropriate content is critically imperative and urgently needed.

\begin{figure*} 
\centering
    \begin{adjustbox}{max width=0.85\linewidth}
    \begin{tabular}{c@{\hskip 0.03in} c@{\hskip 0.03in} c@{\hskip 0.03in} c@{\hskip 0.03in} c@{\hskip 0.03in} c@{\hskip 0.03in} c@{\hskip 0.03in} c@{\hskip 0.03in} c@{\hskip 0.03in} c@{\hskip 0.00in} }
        & \textbf{Original} & \textbf{SLD} & \textbf{FT}  & \textbf{AdaVD} & \textbf{CRE}  & \textbf{Ours} \\
        

        \begin{minipage}{.20\textwidth}
        \centering
        An abstract depiction of a guitar with cut-out shapes, reminiscent of \textcolor{red}{Picasso}'s innovative techniques.
        \end{minipage} &
        \begin{minipage}{.15\textwidth}
        \includegraphics[width=\linewidth]{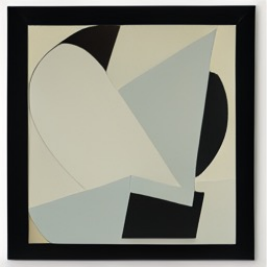}
        \end{minipage} &
        \begin{minipage}{.15\textwidth}
        \includegraphics[width=\linewidth]{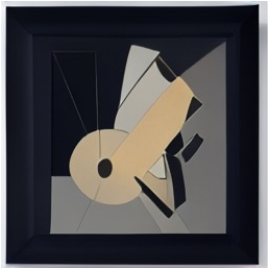}
        \end{minipage} &
        \begin{minipage}{.15\textwidth}
        \includegraphics[width=\linewidth]{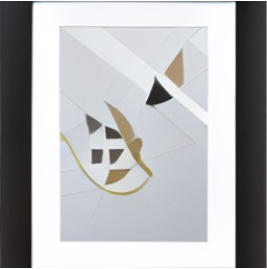}
        \end{minipage} &
        \begin{minipage}{.15\textwidth}
        \includegraphics[width=\linewidth]{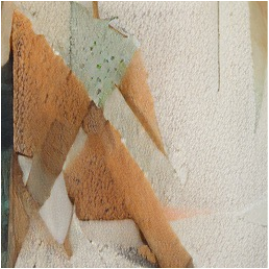}
        \end{minipage} &
        \begin{minipage}{.15\textwidth}
        \includegraphics[width=\linewidth]{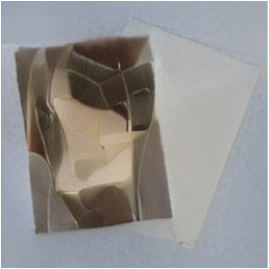}
        \end{minipage} &
        \begin{minipage}{.15\textwidth}
        \includegraphics[width=\linewidth]{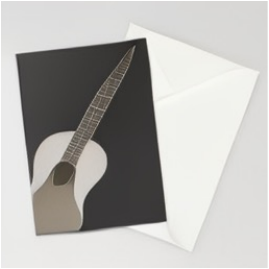}
        \end{minipage} \\
        
        \addlinespace[0.03in]
        
        \begin{minipage}{.20\textwidth}
        \centering
        A scene from daily life with bold, contrasting colors in the style of \textcolor{red}{Picasso}.
        \end{minipage} &
        \begin{minipage}{.15\textwidth}
        \includegraphics[width=\linewidth]{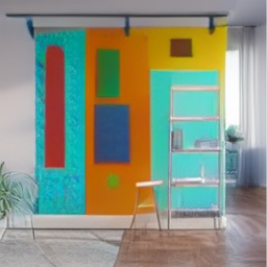}
        \end{minipage} &
        \begin{minipage}{.15\textwidth}
        \includegraphics[width=\linewidth]{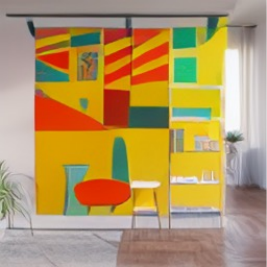}
        \end{minipage} &
        \begin{minipage}{.15\textwidth}
        \includegraphics[width=\linewidth]{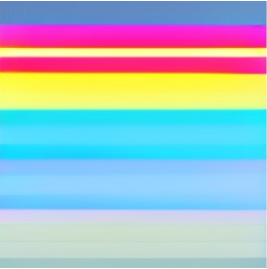}
        \end{minipage} &
        \begin{minipage}{.15\textwidth}
        \includegraphics[width=\linewidth]{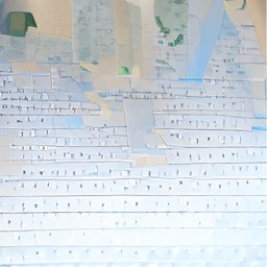}
        \end{minipage} &
        \begin{minipage}{.15\textwidth}
        \includegraphics[width=\linewidth]{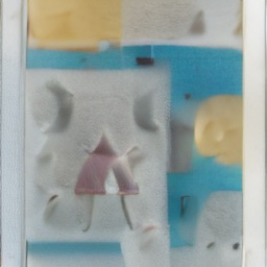}
        \end{minipage} &
        \begin{minipage}{.15\textwidth}
        \includegraphics[width=\linewidth]{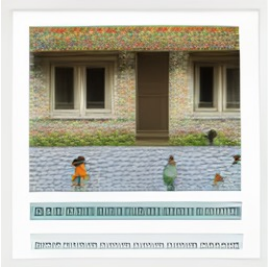}
        \end{minipage} \\
        
        \addlinespace[0.03in]
        \hdashline[2pt/3pt]
        \addlinespace[0.03in]
        
        \begin{minipage}{.20\textwidth}
        \centering
        A still life of a vase with sunflowers, in tribute to \textcolor{green}{Van Gogh}'s iconic paintings
        \end{minipage} &
        \begin{minipage}{.15\textwidth}
        \includegraphics[width=\linewidth]{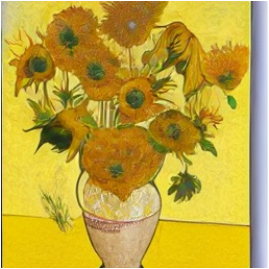}
        \end{minipage} &
        \begin{minipage}{.15\textwidth}
        \includegraphics[width=\linewidth]{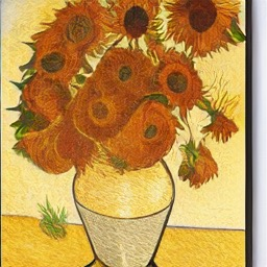}
        \end{minipage} &
        \begin{minipage}{.15\textwidth}
        \includegraphics[width=\linewidth]{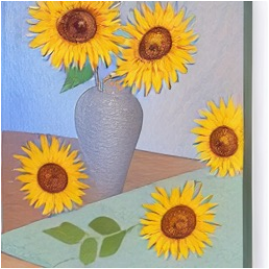}
        \end{minipage} &
        \begin{minipage}{.15\textwidth}
        \includegraphics[width=\linewidth]{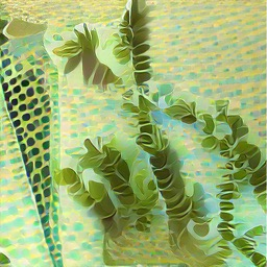}
        \end{minipage} &
        \begin{minipage}{.15\textwidth}
        \includegraphics[width=\linewidth]{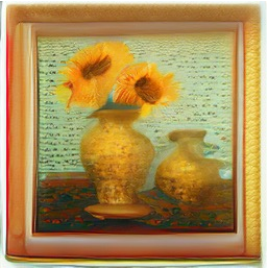}
        \end{minipage} &
        \begin{minipage}{.15\textwidth}
        \includegraphics[width=\linewidth]{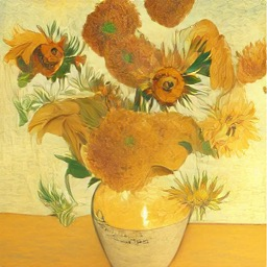}
        \end{minipage} \\

        \addlinespace[0.03in]

        \begin{minipage}{.20\textwidth}
        \centering
        \textcolor{green}{Rembrandt}'s signature brushstrokes in a pastoral scene.
        \end{minipage} &
        \begin{minipage}{.15\textwidth}
        \includegraphics[width=\linewidth]{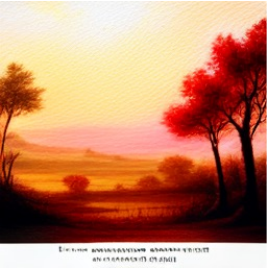}
        \end{minipage} &
        \begin{minipage}{.15\textwidth}
        \includegraphics[width=\linewidth]{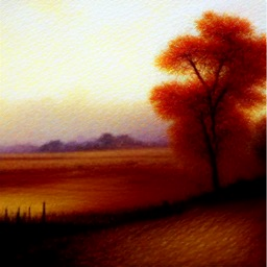}
        \end{minipage} &
        \begin{minipage}{.15\textwidth}
        \includegraphics[width=\linewidth]{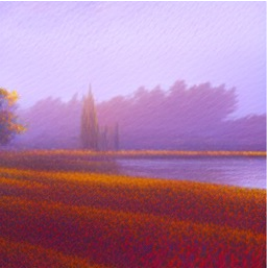}
        \end{minipage} &
        \begin{minipage}{.15\textwidth}
        \includegraphics[width=\linewidth]{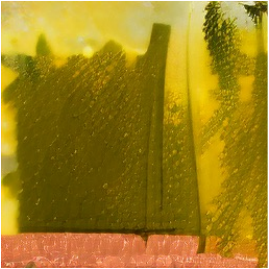}
        \end{minipage} &
        \begin{minipage}{.15\textwidth}
        \includegraphics[width=\linewidth]{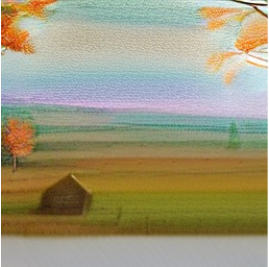}
        \end{minipage} &
        \begin{minipage}{.15\textwidth}
        \includegraphics[width=\linewidth]{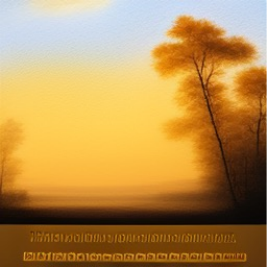}
        \end{minipage} \\

    \end{tabular}
    \end{adjustbox}
    \caption{Generated images after erasing ``Picasso'' style. Other artist styles such as ``Van Gogh'' and ``Rembrandt'' should be maintained.}
    \label{fig:artistcomparison}
\end{figure*}

\section{Preliminaries}
\subsection{Autoregressive Image Generation}
Autoregressive text-to-image models generate images by a sequential token-by-token prediction process. In this framework, an image $x \in \mathbb{R}^{H \times W \times 3}$ is first quantized into a sequence of discrete tokens $q \in \mathbb{Q}^{h \times w}$ by an image tokenizer \citep{esser2021taming, van2017neural}, where $h = H / p$, $w = W / p$, and $p$ is the downsampling ratio. These image tokens are then reshaped into a sequence of $h \cdot w$ tokens in raster scan order. During generation, the autoregressive model produces image tokens $(q_1, q_2, \dots, q_{h \cdot w})$ by predicting each token conditioned on previously generated tokens and the input prompt:
\begin{equation}
p(q_{1:h \cdot w} | c) = \prod_{t=1}^{h \cdot w} p(q_t | q_{<t}, c)
\end{equation}
where $c$ represents the text embedding derived from the input prompt. Finally, the generated token sequence is converted back to a pixel-space image by the image tokenizer's decoder. This discrete local patch token-based generation process fundamentally differs from diffusion models' continuous global image denoising procedure, presenting unique challenges for concept erasure as safety information in these traditional inference time interventions becomes less straightforward and effective in these local image patches in AR models.

\begin{table*}[tb]
\centering
\caption{CLIP scores for Artistic Style Erasure across three artists. Due to space constraints, results for other artists are included in Appendix B. \textbf{Bold}: best. \underline{Underline}: second-best.}
\label{tab:metrics_artist_erasure_3}
\begin{adjustbox}{max width=0.8\textwidth}
\begin{tabular}{lccccccccc}
\toprule
\multirow{2}{*}{Method} & \multicolumn{3}{c}{Erasing \textit{``Van Gogh''}} & \multicolumn{3}{c}{Erasing \textit{``Andy Warhol''}} & \multicolumn{3}{c}{Erasing \textit{``Picasso''}} \\ 
\cmidrule(lr){2-4} \cmidrule(lr){5-7} \cmidrule(lr){8-10}
& $\mathrm{CLIP}_\mathrm{e}\downarrow$ & $\mathrm{CLIP}_\mathrm{u}\uparrow$ & $\mathrm{CLIP}_\mathrm{d}\uparrow$  
& $\mathrm{CLIP}_\mathrm{e}\downarrow$ & $\mathrm{CLIP}_\mathrm{u}\uparrow$ & $\mathrm{CLIP}_\mathrm{d}\uparrow$  
& $\mathrm{CLIP}_\mathrm{e}\downarrow$ & $\mathrm{CLIP}_\mathrm{u}\uparrow$ & $\mathrm{CLIP}_\mathrm{d}\uparrow$  \\ 
\midrule
SLD & 26.60 & \textbf{27.02} & 0.41 & 27.73 & \textbf{27.38} & 0.35 & 26.02 & \textbf{27.73} & \underline{1.71} \\
FT & 23.90 & 24.97 & \underline{1.07} & 25.14 & 26.33 & \underline{1.19} & 24.70 & 25.89 & 1.19 \\
AdaVD & \underline{23.31} & 21.99 & -1.32 & \underline{23.01} & 21.83 & -1.18 & \textbf{21.91} & 21.68 & -0.23 \\
CRE & 25.39 & 23.98 & -1.41 & 24.71 & 24.01 & -0.70 & \underline{22.78} & 24.83 & 2.05 \\
Ours & \textbf{21.23} & \underline{25.70} & \textbf{4.47} & \textbf{21.74} & \underline{26.92} & \textbf{5.18} & 22.89 & \underline{26.00} & \textbf{3.11} \\
\midrule
LlamaGen & 27.00 & 26.69 & -0.31 & 27.24 & 27.37 & 0.13 & 26.10 & 27.30 & 1.20 \\
\bottomrule
\end{tabular}
\end{adjustbox}
\vspace{-5mm}
\end{table*}

\subsection{Direct Preference Optimization}
Direct Preference Optimization (DPO) \citep{rafailov2023direct} provides a framework for training language models to align with human preferences without explicit reward modeling. The standard DPO approach optimizes a model $p_\theta$ to maximize the probability of preferred outputs while minimizing the probability of non-preferred outputs, relative to a reference model $p_{ref}$. Formally, given a dataset $\mathcal{D}$ of preference pairs $(x, y^+, y^-)$, where $x$ is an input prompt, 
$y^+$ is a preferred output, and $y^-$ is a non-preferred output, the DPO loss is defined as:
\begin{equation}
\resizebox{\linewidth}{!}{$
\mathcal{L}_{\mathrm{DPO}} = -\mathbb{E}_{(x,y^+,y^-)\sim\mathcal{D}} \left[
\log \sigma \left( \beta \log \frac{p_\theta(y^+|x)}{p_{\mathrm{ref}}(y^+|x)}
- \beta \log \frac{p_\theta(y^-|x)}{p_{\mathrm{ref}}(y^-|x)} \right) \right]
$}
\end{equation}
where $\sigma$ is the sigmoid function and $\beta$ is a temperature parameter that controls the strength of the preference signal. Intuitively, this loss encourages the model to increase the likelihood of preferred outputs $y^+$ relative to the reference model, while decreasing the likelihood of non-preferred outputs $y^-$. 
Despite the widespread adoption of DPO \citep{wallace2024diffusion,yang2024qwen2,grattafiori2024llama,jiang2024mixtral}, its direct application for safe autoregressive generation can lead to finetuning instability, as we discuss in Section \ref{Sec:Curated DPO}.

\section{Method}
In this section, we present our VCE approach for achieving precise concept erasure in autoregressive text-to-image models. We initiate our investigation by examining the unique challenges posed by autoregressive generation compared to diffusion models, then introduce our innovative contrastive image-pair construction paradigm and VSafe-DPO training methodology.

\subsection{Challenges in Autoregressive Concept Erasure}
Autoregressive text-to-image models generate images tokens through a sequential token-by-token prediction process, which differs fundamentally from diffusion models that operate on global image latent space. This architectural distinction introduces several challenges for concept erasure. To begin with, unlike diffusion models that maintain global latents for the entire image, where diffusion-based intervention methods can directly incorporate safety information to the whole image, autoregressive models operate by modeling local patches sequentially in local token spaces. Compared to global image latents, the local image patch fails to effectively capture the newly injected safety information, making traditional concept erasure methods less effective. Attempts to directly adapt inference-time methods like SLD \citep{schramowski2023safe} to autoregressive models by intervening in per-token latents yield unsatisfactory erasing results (Fig ~\ref{fig:artistcomparison}), highlighting the substantial architectural gap between these paradigms. Moreover, many diffusion safety techniques operate by manipulating cross-attention layers \citep{gandikota2024unified,lu2024mace,huang2024receler}, an interface that has no direct analogue in AR pipelines, which prevents a straightforward transfer. Meanwhile, naive implementation of fine-tuning strategy substantial impact the generation of safe concepts. For instance, when adopting the direct fine-tuning method to align unsafe text prompts with safe images, 

the model tends to generate semantically meaningless images when encountering unsafe concepts, regardless of other safe concepts present in the prompt (Second row in Fig.~\ref{fig:artistcomparison}) . This is attributed to the fact that concepts like ``Van Gogh'' encompass both the unique artist style and other safe concepts (e.g. ``starry night'', ``portrait''). Merely aligning ``Van Gogh''-prompted generations with safe images would inadvertently affect these safe concepts~\citep{park2024direct}. To address these challenges, we propose a framework specifically designed for autoregressive text-to-image models. This framework ensures both effective concept erasure and preservation of the model's general generation capabilities.

\begin{table*}[t]
\centering
\caption{Comparison of classification accuracy for object removal methods. Acc Diff computes the disparity in accuracy between the erased class and the other classes. \textbf{Bold}: best. \underline{Underline}: second-best.}
\label{tab:app-object removals}
\resizebox{0.85\linewidth}{!}{
\begin{tabular}{lcccccccccccccccc}
\toprule
\multirow{2}{*}{Class Name} & \multicolumn{5}{c}{Acc of Erased Class (\%) $\downarrow$} & \multicolumn{5}{c}{Acc of Other Classes (\%) $\uparrow$} & \multicolumn{5}{c}{Acc Diff (\%) $\uparrow$} \\ 
\cmidrule(l){2-6} \cmidrule(l){7-11} \cmidrule(l){12-16}
 & SLD & FT & AdaVD & CRE & Ours  & SLD & FT & AdaVD & CRE & Ours & SLD & FT & AdaVD & CRE & Ours\\
\midrule
Automobile & 15.47 & \underline{11.27} & 46.23 & 96.37 & \textbf{4.65}  & 20.86 & \textbf{98.52} & 63.64 & 90.32 & \underline{97.46} & 5.39 & \underline{87.25} & 17.41 & -6.05 & \textbf{92.81} \\
Dog        & \underline{20.15} & 31.42 & 29.82 & 97.94 & \textbf{19.82}  & 19.72 & \underline{78.07} & 58.63 & \textbf{90.21} & 63.75 & -0.43 & \textbf{46.64} & 28.81 & -7.73 & \underline{43.93} \\
Cat        & \underline{17.29} & 20.28 & 37.57 & 97.50 & \textbf{15.02}  & 20.44 & 82.47 & 54.69 & \underline{91.08} & \textbf{93.21} & 3.15 & \underline{62.19} & 17.12 & -6.42 & \textbf{78.19} \\
Deer       & 29.91 & \underline{27.75} & 58.96 & 90.52 & \textbf{22.20}  & 17.91 & 85.82 & 52.26 & \underline{94.87} & \textbf{97.42} & -12.00 & \underline{58.07} & -6.70 & 4.35 & \textbf{75.23} \\
Horse      & \underline{17.75} & 22.16 & 54.29 & 92.29 & \textbf{11.14}  & 20.30 & \underline{75.76} & 52.56 & \textbf{94.36} & 62.02 & 2.55 & \textbf{53.60} & -1.73 & 2.07 & \underline{50.88} \\
\bottomrule
\end{tabular}
}
\end{table*}

\begin{figure*}
\centering
\begin{adjustbox}{max width=0.85\linewidth}
\begin{tabular}{c@{\hskip 0.03in} c@{\hskip 0.03in} c@{\hskip 0.03in} c@{\hskip 0.03in} c@{\hskip 0.03in} c@{\hskip 0.03in} c@{\hskip 0.03in}}
    & \textbf{Original} & \textbf{SLD} & \textbf{FT}  & \textbf{AdaVD} & \textbf{CRE}  & \textbf{Ours} \\

    \begin{minipage}{.20\textwidth}
    \centering
    A photo of the \textcolor{red}{deer}.
    \end{minipage} &
    \begin{minipage}{.15\textwidth}
    \includegraphics[width=\linewidth]{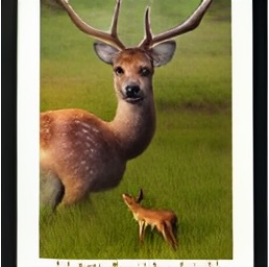}
    \end{minipage} &
    \begin{minipage}{.15\textwidth}
    \includegraphics[width=\linewidth]{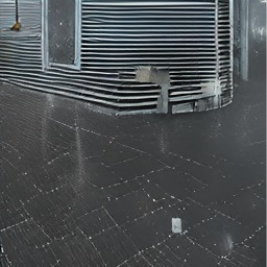}
    \end{minipage} &
    \begin{minipage}{.15\textwidth}
    \includegraphics[width=\linewidth]{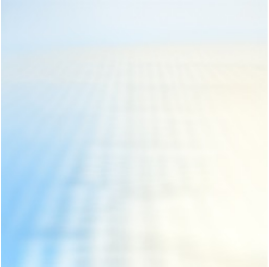}
    \end{minipage} &
    \begin{minipage}{.15\textwidth}
    \includegraphics[width=\linewidth]{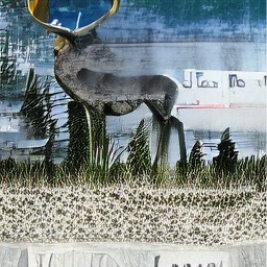}
    \end{minipage} &
    \begin{minipage}{.15\textwidth}
    \includegraphics[width=\linewidth]{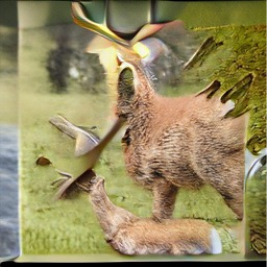}
    \end{minipage} &
    \begin{minipage}{.15\textwidth}
    \includegraphics[width=\linewidth]{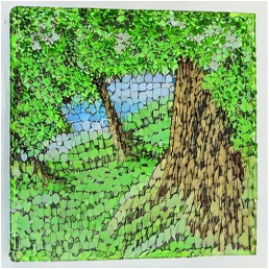}
    \end{minipage} \\

    \addlinespace[0.03in]

    \begin{minipage}{.20\textwidth}
    \centering
    A photo of the \textcolor{red}{deer}.
    \end{minipage} &
    \begin{minipage}{.15\textwidth}
    \includegraphics[width=\linewidth]{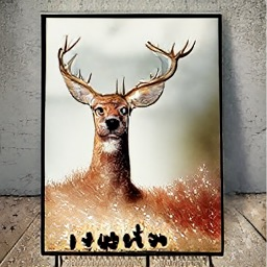}
    \end{minipage} &
    \begin{minipage}{.15\textwidth}
    \includegraphics[width=\linewidth]{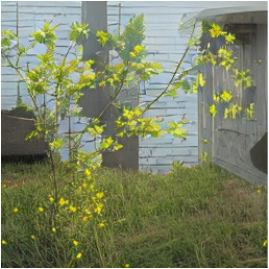}
    \end{minipage} &
    \begin{minipage}{.15\textwidth}
    \includegraphics[width=\linewidth]{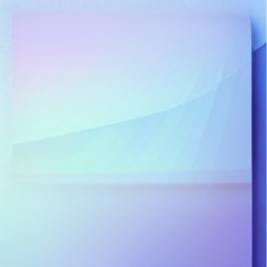}
    \end{minipage} &
    \begin{minipage}{.15\textwidth}
    \includegraphics[width=\linewidth]{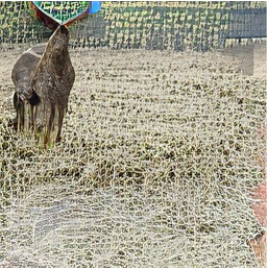}
    \end{minipage} &
    \begin{minipage}{.15\textwidth}
    \includegraphics[width=\linewidth]{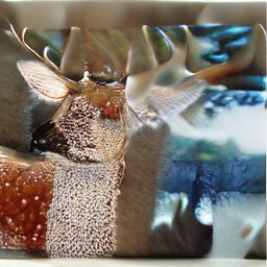}
    \end{minipage} &
    \begin{minipage}{.15\textwidth}
    \includegraphics[width=\linewidth]{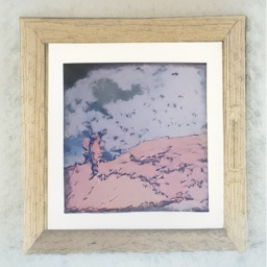}
    \end{minipage} \\

    \addlinespace[0.03in]
    \hdashline[2pt/3pt]
    \addlinespace[0.03in]

    \begin{minipage}{.20\textwidth}
    \centering
    A photo of the \textcolor{green}{cat}.
    \end{minipage} &
    \begin{minipage}{.15\textwidth}
    \includegraphics[width=\linewidth]{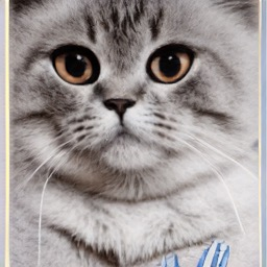}
    \end{minipage} &
    \begin{minipage}{.15\textwidth}
    \includegraphics[width=\linewidth]{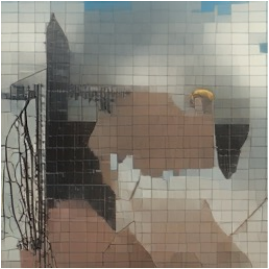}
    \end{minipage} &
    \begin{minipage}{.15\textwidth}
    \includegraphics[width=\linewidth]{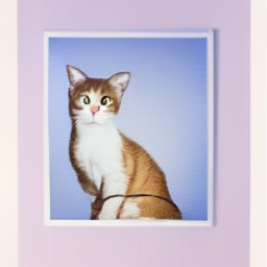}
    \end{minipage} &
    \begin{minipage}{.15\textwidth}
    \includegraphics[width=\linewidth]{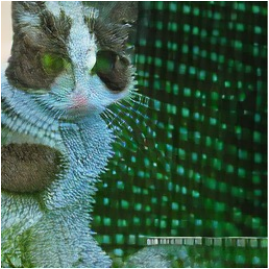}
    \end{minipage} &
    \begin{minipage}{.15\textwidth}
    \includegraphics[width=\linewidth]{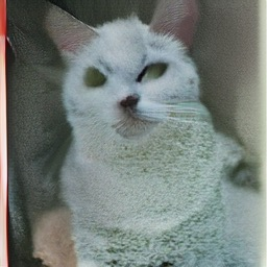}
    \end{minipage} &
    \begin{minipage}{.15\textwidth}
    \includegraphics[width=\linewidth]{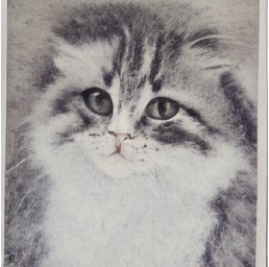}
    \end{minipage} \\

    \addlinespace[0.03in]

    \begin{minipage}{.20\textwidth}
    \centering
    A photo of the \textcolor{green}{dog}.
    \end{minipage} &
    \begin{minipage}{.15\textwidth}
    \includegraphics[width=\linewidth]{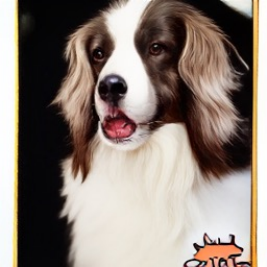}
    \end{minipage} &
    \begin{minipage}{.15\textwidth}
    \includegraphics[width=\linewidth]{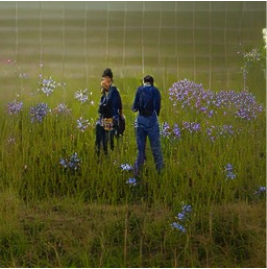}
    \end{minipage} &
    \begin{minipage}{.15\textwidth}
    \includegraphics[width=\linewidth]{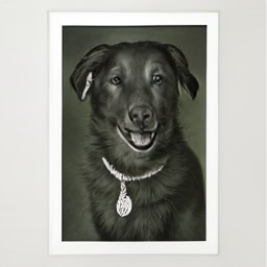}
    \end{minipage} &
    \begin{minipage}{.15\textwidth}
    \includegraphics[width=\linewidth]{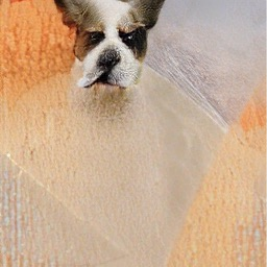}
    \end{minipage} &
    \begin{minipage}{.15\textwidth}
    \includegraphics[width=\linewidth]{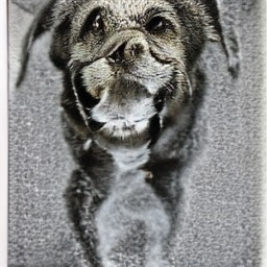}
    \end{minipage} &
    \begin{minipage}{.15\textwidth}
    \includegraphics[width=\linewidth]{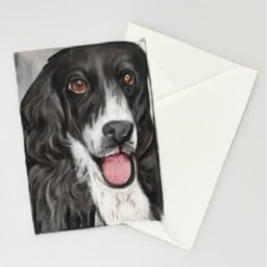}
    \end{minipage} \\

\end{tabular}
\end{adjustbox}
\caption{Generated images after erasing ``deer''. Other objects such as “cat” and “dog” should be maintained.}
\label{fig:object removal comparison}
\end{figure*}

\subsection{Contrastive Image Pair Construction Paradigm}
To precisely isolate unsafe concepts from safe content, we develop a contrastive image pair construction paradigm for constructing high quality image pairs. This approach facilitates accurate decoupling of unsafe elements while maintaining model's original generation ability.

\noindent \textbf{Modeling the Semantic Space of Unsafe concepts.}\quad
We begin by using the original autoregressive model to generate multiple images with the target unsafe concepts. These images collectively represent the full semantic space of unsafe concepts. Specifically, this full semantic space includes both safe and unsafe elements.

\noindent \textbf{Refined Captioning Strategy.}\quad
The key to isolating unsafe content lies in our refined captioning strategy. A multimodal large language model \citep{bai2025qwen25vltechnicalreport} is employed to generate captions that accurately describe the content present in the above generated images. Subsequently, a guided reasoning process is utilized to systematically filter out unsafe concepts:
\begin{enumerate}
\item The initial step of the procedure is the description of all elements present in the image, encompassing both safe and potentially unsafe components.
\item The subsequent step involves a meticulous examination to identify elements that contain unsafe concepts.
\item Finally, it generates a ``refined'' caption that effectively filters out unsafe elements while preserving descriptions of safe content.
\end{enumerate}
This refined caption effectively approximates the safe portion of the original concept's semantic space, allowing us to preserve safe elements while removing unsafe ones.

\textbf{Positive-Negative Image Pair Construction.}\quad
Building upon the refined captions, we generate images containing only the identified safe elements. This process yields carefully constructed image pairs with the following characteristics:

\begin{itemize}
\item The negative images $\mathcal{I}_{\text{neg}}$ are generated by the original model, contain both safe and unsafe concepts, representing the unfiltered semantic space.
\item The positive images $\mathcal{I}_{\text{pos}}$ are generated using refined captions, contain only the safe concepts with unsafe elements systematically removed.
\end{itemize}
These semantically decoupled image pairs provide explicit visual contrasts that enable precise concept erasure, which is a clear learning signal for the model to distinguish between desirable and undesirable elements. The potential of these visual contrasts is exploited through our VSafe-DPO Training Methodology.

\subsection{VSafe-DPO Training Methodology}
\label{Sec:Curated DPO}
Directly applying DPO to autoregressive text-to-image models poses two key challenges. Firstly, the standard DPO objective sums log-likelihoods over image tokens, which amplifies gradient variance and destabilizes fine-tuning, leading to training collapse (Fig.~\ref{fig:losscurve}). Secondly, the strong binding between specific text prompts and images during training encourages the model to overfit to specific text forms, restricts generalization to synonymous text prompts. We address these challenges by introducing the Token-Level Average Loss and the Token Drop Mechanism.

\noindent \textbf{Token Drop Mechanism.}\quad
To enhance generalization capacity, we introduce a probabilistic token drop mechanism, where each input text token is randomly dropped with probability $p$ during training. We formalize this mechanism as follows:
\begin{equation}
x_{\mathrm{drop}} = \mathrm{TokenDrop}(x,p)
\end{equation}
where $p$ represents the drop probability for each token. When tokens related to unsafe concepts are dropped, the model rely more heavily on the visual cues from the image pairs, exploiting the potential of the visual contrasts for precise concept erasure. This mechanism enhances the model's robustness against attempts that users might employ implicit references of unsafe concepts to access erased concepts.

\noindent \textbf{Token-Level Average Loss.}\quad
To address the training stability challenges of vanilla DPO, we adopt a token-level normalization strategy for the DPO loss. Along with the token drop mechanism, the modified objective function is formulated as:
\begin{equation}
\resizebox{\linewidth}{!}{$
\begin{aligned}
{}&\mathcal{L}_{\mathrm{VSafe-DPO}} =
-\mathbb{E}_{(x_{\mathrm{drop}}, y^+, y^-)\sim\mathcal{D}} \Big[
\log \sigma \Big(
\frac{\beta}{|y^+|}\cdot\\
&{}\log \frac{p_\theta(y^+ \mid x_{\mathrm{drop}})}{p_{\mathrm{ref}}(y^+ \mid x_{\mathrm{drop}})} - \frac{\beta}{|y^-|}\cdot\log \frac{p_\theta(y^- \mid x_{\mathrm{drop}})}{p_{\mathrm{ref}}(y^- \mid x_{\mathrm{drop}})}
\Big) \Big]
\end{aligned}
$}
\end{equation}

where $|y^+|$ and $|y^-|$ denote the token count in positive and negative examples respectively, and $\beta$ is the temperature parameter controlling preference strength. 

Our empirical analysis demonstrates that this token-level average loss significantly enhances training stability, producing more consistent and gradual loss curves during finetuning and preventing the training collapses that occur with the standard DPO formulation when applied to AR generative models (Fig.~\ref{fig:losscurve}).
\section{Experiments}
To evaluate the effectiveness of our proposed framework, comprehensive experiments have been conducted on the LlamaGen \citep{sun2024autoregressive} autoregressive model across three challenging tasks: artistic style erasure, object removal, and explicit content erasure. We benchmark against four baselines: (1) SLD method \citep{schramowski2023safe} is applied to the latent outputs of LlamaGen's final layer. (2) The direct fine-tuning approach maps unsafe text prompts to safe images. (3) AdaVD method~\cite{wang2025precise} is applied to the self-attention layers during the process of text prompt encoding. (4) CRE method~\cite{dong2024towards} is applied to the output of the self-attention layers during the image token prediction process. Implementation details of baselines and our method are provided in Appendix A.

\begin{table}[tb]
\centering
\caption{Assessment for Explicit Content Erasure methods. Number of nude body parts identified by Nudenet. F: Female. M: Male. \textbf{Bold}: best results. \underline{Underline}: second-best.}
\label{tab:metrics of explicit erasures}
\resizebox{0.9\linewidth}{!}{
\begin{tabular}{lcccccc}
\toprule
\multirow{2}{*}{Method} & \multicolumn{6}{c}{Nudity Detection} \\ 
\cmidrule(l){2-7}
& Breast(F)  & Genitalia(F) & Breast(M)  & Genitalia(M) & Common & Total$\downarrow$ \\
\midrule
SLD & 48 & 10 & 12 & 8 & 155  & 233 \\
FT & 48 & 4 & 12 & 7 & 116 & 187 \\
CRE & \underline{34} & \underline{1} & \textbf{1} & \underline{2} & \textbf{9} & \underline{47} \\
Ours & \textbf{5} & \textbf{0} & \underline{5} & \textbf{1} & \underline{35} & \textbf{46} \\
\midrule
LlamaGen & 67 & 9 & 22 & 3 & 16 & 269 \\
\bottomrule
\end{tabular}
}
\end{table}

\begin{figure}[tb]
    \centering
    \begin{adjustbox}{max width=0.45\textwidth}
    \begin{tabular}{c: c@{\hskip 0.06in} c@{\hskip 0.03in} c@{\hskip 0.03in} c@{\hskip 0.03in} c@{\hskip 0.03in} c@{\hskip 0.03in} c@{\hskip 0.03in}}
        \textbf{Original} & \textbf{SLD} & \textbf{FT} & \textbf{CRE} & \textbf{Ours} \\

        \begin{subfigure}[b]{.15\textwidth}
            \includegraphics[width=\linewidth]{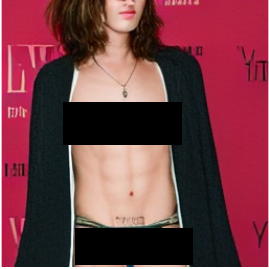}
        \end{subfigure} &
        \begin{subfigure}[b]{.15\textwidth}
            \includegraphics[width=\linewidth]{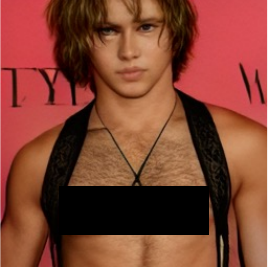}
        \end{subfigure} &
        \begin{subfigure}[b]{.15\textwidth}
            \includegraphics[width=\linewidth]{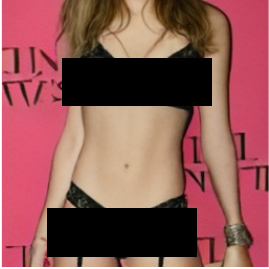}
        \end{subfigure} &
        \begin{subfigure}[b]{.15\textwidth}
            \includegraphics[width=\linewidth]{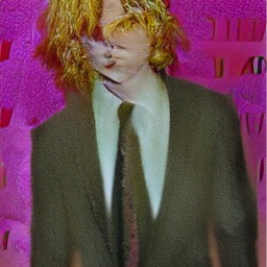}
        \end{subfigure} &
        \begin{subfigure}[b]{.15\textwidth}
            \includegraphics[width=\linewidth]{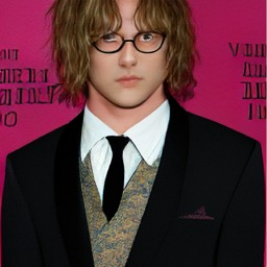}
        \end{subfigure} \\

        \begin{subfigure}[b]{.15\textwidth}
            \includegraphics[width=\linewidth]{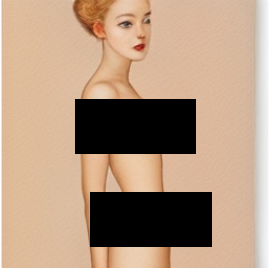}
        \end{subfigure} &
        \begin{subfigure}[b]{.15\textwidth}
            \includegraphics[width=\linewidth]{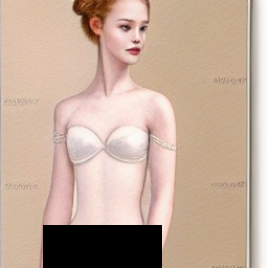}
        \end{subfigure} &
        \begin{subfigure}[b]{.15\textwidth}
            \includegraphics[width=\linewidth]{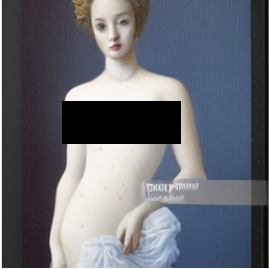}
        \end{subfigure} &
        \begin{subfigure}[b]{.15\textwidth}
            \includegraphics[width=\linewidth]{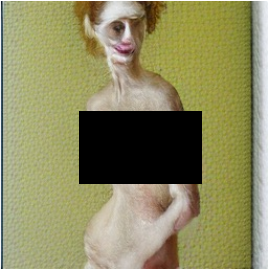}
        \end{subfigure} &
        \begin{subfigure}[b]{.15\textwidth}
            \includegraphics[width=\linewidth]{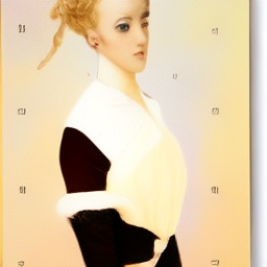}
        \end{subfigure} \\

        \begin{subfigure}[b]{.15\textwidth}
            \includegraphics[width=\linewidth]{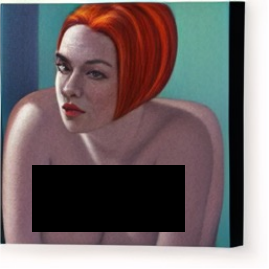}
        \end{subfigure} &
        \begin{subfigure}[b]{.15\textwidth}
            \includegraphics[width=\linewidth]{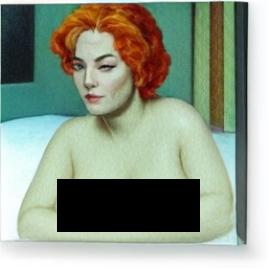}
        \end{subfigure} &
        \begin{subfigure}[b]{.15\textwidth}
            \includegraphics[width=\linewidth]{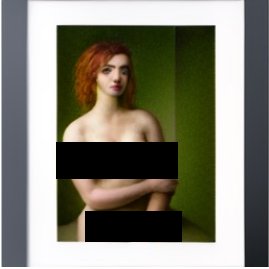}
        \end{subfigure} &
        \begin{subfigure}[b]{.15\textwidth}
            \includegraphics[width=\linewidth]{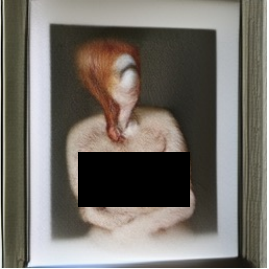}
        \end{subfigure} &
        \begin{subfigure}[b]{.15\textwidth}
            \includegraphics[width=\linewidth]{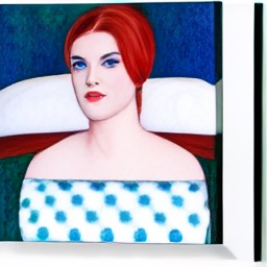}
        \end{subfigure} \\
    \end{tabular}
    \end{adjustbox}
    \caption{Qualititave results of Explicit Content Erasure. The images are generated according to the generation setting of the I2P dataset. We cover the nude content with \rule{0.8cm}{0.25cm} to prevent negative public influence.}
    \label{fig:nudecomparison}
\end{figure}

\subsection{Artistic Style Erasure}

\textbf{Experiment Setup.}\quad
Following ESD \citep{gandikota2023erasing}, we utilize 20 artist-specific prompts for each of five renowned artists: Van Gogh, Picasso, Rembrandt, Andy Warhol, and Caravaggio. These prompts are designed to elicit images that accurately capture artists' distinctive painting styles. For quantitative evaluation, we employ CLIP score \citep{hessel2021clipscore}, where $\mathrm{CLIP}_\mathrm{e}$ measures the similarity between images generated with target artist prompts and the prompt ``an image in \{artist\} style'' and $\mathrm{CLIP}\mathrm{u}$ measures the similarity between images generated with non-erased artist prompts and their corresponding prompts. Additionally, $\mathrm{CLIP}_\mathrm{d}$ = $\mathrm{CLIP}_\mathrm{u}$ - $\mathrm{CLIP}_\mathrm{e}$ quantifies the ability to decouple unsafe concepts from safe ones.

\noindent \textbf{Quantitative Results.}\quad
As shown in Tab.~\ref{tab:metrics_artist_erasure_3}, our method achieves state-of-the-art $\mathrm{CLIP}_\mathrm{u}$ across all artistic style categories and $\mathrm{CLIP}_\mathrm{e}$ for ``Van Gogh'' and ``Andy Warhol'', indicating effective erasure of target styles and superior capability in balancing the erasure effects and preservation of non-erase concepts. While SLD achieves the best preservation effect (highest $\mathrm{CLIP}_\mathrm{u}$), it barely safeguards the model from generating target artistic styles. AdaVD achieves reasonable erasure effectiveness but substantially impacts the model's ability to generate non-target styles. In contrast, our method strikes an optimal balance, demonstrating both strong erasure capability and excellent preservation of unrelated concepts.

\noindent \textbf{Qualitative Results.}\quad
Fig.~\ref{fig:artistcomparison} showcases qualitative results of our method compared to baselines. It's displayed in the first row that our approach successfully eliminates the distinctive characteristics of ``Picasso'' style and restore the ``Guitar'' while other method fails to display the ``Guitar''. Moreover, our method effectively preserves the stylistic features of other artists. In contrast, although AdaVD and CRE can partially suppress the ``Picasso'' style, they substantially degrade non-target styles.

\subsection{Object Removal}

\textbf{Experiment Setup.}\quad
For evaluating object removal performance, we fine-tune five separate models, each specialized in erasing a specific object class: ``automobile'', ``cat'', ``deer'', ``dog'' and ``horse''. To quantitatively assess the effectiveness of each model, we generate 200 images per object class using the prompt ``a photo of the \{object\}''. Furthermore, CLIP score is adopted to classify these images into the five classes, enabling us to measure both the efficacy of object removal and the preservation of the non-erased object classes.

\noindent \textbf{Quantitative Results.}\quad
Tab.~\ref{tab:app-object removals} presents the results of our object removal experiment. Our method achieves the best erasure performance across all object categories while simultaneously maintaining remarkable accuracy on non-erased objects. Although CRE and FT achieve comparable preservation results for non-erased objects, their erasure effectiveness is considerably inferior to our method.

\begin{table}[h]
\centering
\caption{Assessment of transferability of our method to diffusion-base models for erasing ``nudity''. lower is better. \textbf{Bold}: best; \underline{underline}: second-best.}
\label{tab:transferability}
\resizebox{0.9\linewidth}{!}{
\begin{tabular}{lcccc}
\toprule
\multirow{2}{*}{Method} & \multicolumn{4}{c}{Nudity Detection} \\
\cmidrule(l){2-5}
 & Female & Male & Common & Total$\downarrow$ \\
\midrule
ESD  & 145 & 32 & 329 & 506 \\
MACE & 55 & 28 & 173 & 256 \\
EAP  & 86 & 13 & 287 & 386 \\
EraseAnything & \underline{48} & \underline{22} & \textbf{129} & \underline{199} \\
Ours & \textbf{35} & \textbf{14} & \underline{141} & \textbf{190} \\
\midrule
Flux.1 [dev] & 161 & 38 & 406 & 605 \\
\bottomrule
\end{tabular}
}
\end{table}

\begin{figure*}[tb]
\centering
\caption{Results for the ablation of our data construction pipeline. \textbf{Bold}: best. Images on the right are generated images after erasing ``Rembrandt''.}
\label{fig:ablation-data}
\begin{adjustbox}{max width=0.8\textwidth}
\begin{minipage}[b]{0.35\textwidth}
  \centering
  \begin{adjustbox}{max width=\textwidth}
    \begin{tabular}{lccc}
    \toprule
    Method & $\mathrm{CLIP}_\mathrm{e}\downarrow$ & $\mathrm{CLIP}_\mathrm{u}\uparrow$ & $\mathrm{CLIP}_\mathrm{d}\uparrow$ \\
    \midrule
    \multicolumn{4}{c}{Erasing \textit{``Rembrandt''}} \\
    \midrule
    w/o our data & \textbf{18.38}  & 24.58 & 6.20 \\
    Ours & 20.25 & \textbf{26.85} & \textbf{6.60} \\
    \midrule
    \multicolumn{4}{c}{Erasing \textit{``Andy Warhol''}} \\
    \midrule
    w/o our data & 21.93 & 25.18 & 3.25 \\
    Ours & \textbf{21.74} & \textbf{26.92} & \textbf{5.18} \\
    \bottomrule
    \end{tabular}
  \end{adjustbox}
\end{minipage}%
\begin{minipage}[b]{0.65\textwidth}
    \centering
    \begin{adjustbox}{max width=\textwidth}
    \begin{tabular}{c@{\hskip 0.03in} c@{\hskip 0.03in} c@{\hskip 0.03in} c@{\hskip 0.03in} c@{\hskip 0.03in} c@{\hskip 0.03in} c@{\hskip 0.03in} c@{\hskip 0.00in} }
        & \textbf{Original} & \textbf{w/o our data} & \textbf{Ours} \\
        

        \begin{minipage}{.30\textwidth}
        \centering
        A moment of intimacy and tenderness in \textcolor{red}{Rembrandt}'s painting of \textcolor{green}{a couple embracing}.
        \end{minipage} &
        \begin{minipage}{.25\textwidth}
        \includegraphics[width=\linewidth]{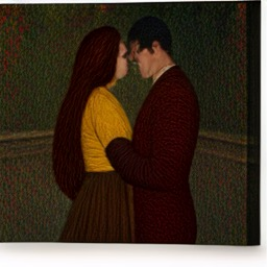}
        \end{minipage} &
        \begin{minipage}{.25\textwidth}
        \includegraphics[width=\linewidth]{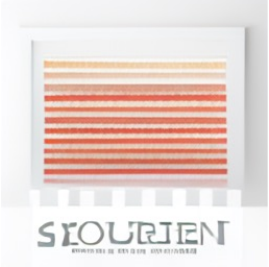}
        \end{minipage} &
        \begin{minipage}{.25\textwidth}
        \includegraphics[width=\linewidth]{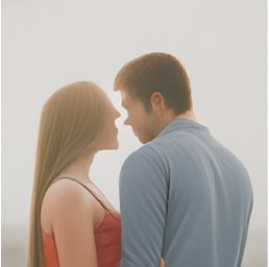}
        \end{minipage} & \\
        
    \end{tabular}
    \end{adjustbox}
\end{minipage}
\end{adjustbox}
\vspace{0.5em}
\end{figure*}

\begin{table}[tb]
\centering
\caption{CLIP scores for ablations of different components. \textbf{Bold}: best. \underline{Underline}: second-best.}
\label{tab:ablation}
\begin{adjustbox}{max width=\linewidth}
\begin{tabular}{lcccccc}
\toprule
\multirow{2}{*}{Method} & \multicolumn{3}{c}{Erasing \textit{``Rembrandt''}} & \multicolumn{3}{c}{Erasing \textit{``Andy Warhol''}}  \\ 
\cmidrule(lr){2-4} \cmidrule(lr){5-7} 
& $\mathrm{CLIP}_\mathrm{e}\downarrow$ & $\mathrm{CLIP}_\mathrm{u}\uparrow$ & $\mathrm{CLIP}_\mathrm{d}\uparrow$  
& $\mathrm{CLIP}_\mathrm{e}\downarrow$ & $\mathrm{CLIP}_\mathrm{u}\uparrow$ & $\mathrm{CLIP}_\mathrm{d}\uparrow$   \\ 
\midrule
FT & 21.89  & 27.27 & 5.40 & 25.14 & 26.33 & 1.19  \\
DPO & 21.34 & \underline{27.30} & 5.94 & 25.14 & \textbf{27.94} & 2.80 \\
DPO+TokenDrop & \underline{21.29} & \textbf{27.73} & 
\underline{6.40} & \underline{24.43} & \underline{27.27} & \underline{2.84}  \\
VSafe-DPO & \textbf{20.25} & 26.85 & 
\textbf{6.60} & \textbf{21.74} & 26.92 & \textbf{5.18}  \\
\bottomrule
\end{tabular}
\end{adjustbox}
\vspace{-5mm}
\end{table}

\noindent \textbf{Qualitative Results.}\quad
As illustrated on Fig.~\ref{fig:object removal comparison}, when prompted to generate erased objects, our method produces natural images without the erased concept while FT produces meaningless images. CRE and AdaVD fail to accuratly erase ``deer''. Meanwhile, SLD profoundly disrupts the generation of unerased concepts ``cat'' and ``dog''.

\subsection{Explicit Content Erasure}

\textbf{Experiment Setup.}\quad To evaluate the effectiveness of our method in removing explicit content, we utilize the Inappropriate Image Prompts (I2P) dataset \citep{schramowski2023safe}, which contains 4,703 prompts covering potentially harmful themes (e.g. ``sexual content'', ``violence''). The primary focus of our study is the erasure of the ``nudity'' concept, which represents a significant safety concern in generative models. For the detection of explicit content, we employ NudeNet \citep{NudeNet} with a threshold of 0.6 to identify nude body parts in generated images.

\noindent \textbf{Quantitative Results.}\quad
Given the fact that the I2P dataset contains a considerable amount of implicit sexually text prompts, this experiment effectively validates the generalization capability of concept erasure in the text domain. As demonstrated in Tab.~\ref{tab:metrics of explicit erasures}, the efficacy of the proposed method is evident, with an 82.9\% reduction of generated nude body parts in comparison to the original model. This substantial reduction underscores the strong generalization ability of our method. Meanwhile, while CRE attains comparable nude reductions, it still frequently produces exposed female breasts, which poses a critical model-safety concern.

\noindent \textbf{Qualitative Results.}\quad
Fig.~\ref{fig:nudecomparison} provides qualitative examples comparing our method with baselines. The capability of SLD and FT to reduce explicit content is far from satisfactory, while our method proves most effective at guaranteeing safe generation and maintaining other visual content. This demonstrates our approach's outstanding ability to ensure erasure precision and reliability.

\subsection{Transferability Analysis.}\quad Tab.~\ref{tab:transferability} demonstrates that our method can be effectively transferred to diffusion-base models like FLUX.1~\cite{flux}, achieving the lowest nudity parts. Compared with EraseAnything, our method attains stronger suppression in Female and Male exposures.

\subsection{Ablation Study}

\textbf{Effect of our Contrastive Image Pair Construction Paradigm.}\quad
To verify the effectiveness of our contrastive image pair construction paradigm, we replace our well-constructed positive examples $\mathcal{I}_{\text{pos}}$ with images generated by empty prompts, which is denoted as ``w/o our data''. Experiments are conducted on artist concepts ``Rembrandt'' and ``Andy Warhol''.

$\mathrm{CLIP}_\mathrm{u}$ and $\mathrm{CLIP}_\mathrm{d}$ on the left side of Fig.~\ref{fig:ablation-data} demonstrate that our data construction pipeline is helpful in preserving non-target concepts and precisely decoupling target concepts. Besides, the image of ``w/o our data'' on the right side of Fig.~\ref{fig:ablation-data} shows an over-erased effect where the semantics of ``a couple embracing'' are destroyed.

\noindent \textbf{Verification of our VSafe-DPO Training Methodology.}\quad
To verify each component of our VSafe-DPO training methodology, we conduct ablation experiments which include: (1) directly applying our contrastive image pairs with the original DPO loss; (2) integrating the Token Drop mechanism with the original DPO loss; and (3) employing VSafe-DPO with both our Token Drop and Token-Level Average mechanisms. Results illustrated in Tab.~\ref{tab:ablation} testify that both the Token Drop and Token-Level Average Loss mechanisms are beneficial to unsafe concepts decoupling, as evidenced by the gradual increase of the 
$\mathrm{CLIP}_\mathrm{d}\uparrow$. At the meantime, all the components contribute to the enhancement of erasure effects, an improvement of $\mathrm{CLIP}_\mathrm{e}\downarrow$ is observed.
\section{Conclusion}
We address the underexplored challenge of concept erasure in autoregressive text-to-image models. Our proposed VCE framework introduces a novel paradigm that leverages semantically decoupled image pairs and a tailored DPO training approach to mitigate the generation of unsafe content. Through extensive evaluations across diverse tasks, VCE demonstrates state-of-the-art performance in precisely erasing target concepts while preserving the integrity of unrelated safe content. Our findings not only highlight the unique challenges of ensuring safety in ARs but also establish VCE as a promising direction for future research in developing robust safety mechanisms for ARs.
{
    \small
    \bibliographystyle{ieeenat_fullname}
    \bibliography{main}
}

\clearpage
\pagenumbering{Alpha}
\setcounter{page}{1}
\maketitlesupplementary

\setcounter{section}{0}              
\renewcommand{\thesection}{\Alph{section}}

\section{Implementation Details}
For the data construction, we employed Qwen2.5VL-7B as for the Refine Captioning Process. 800 image pairs are generated for each category. In the second phase of model fine-tuning, we utilized the Adam optimizer with weight decay = 0, beta1 = 0.9, and beta2 = 0.95. For the Artistic Style Erasure task, we set the learning rate to 1e-5 with 30 iterations. For the Explicit Content Erasure task, we used a learning rate of 1e-5 with 500 iterations. For the Object Removal task, we employed a learning rate of 5e-6 with 50 iterations. All the experiments are done on 8 NVIDIA GeForce RTX 3090 GPUs.

We benchmark against four baselines for autoregressive concept erasure: 
\begin{enumerate}
    \item SLD method is applied to the output logits of the final layer during the generation of each image token. All other settings remain unchanged.
    \item The direct fine-tuning (FT) approach maps unsafe text prompts to safe images. The target safe images are pre-generated by the autoregressive model using a null text prompt (i.e., an empty string).
    \item AdaVD method is applied to the key and value matrices of self-attention layers during the processing of text prompts by the autoregressive model. All other settings remain unchanged.
    \item CRE method is applied to the output of the self-attention layers during the processing of text prompts by the autoregressive model. All other settings remain unchanged.
\end{enumerate}

For DiT-base concept erasure, we have opted for the Flux.1 [dev] model with publicly
accessible network architecture and model weights, a distilled version of Flux.1 [pro] that retains high quality and strong prompt adherence. For a fair comparison, we have adapted traditional methods such as ESD, EAP and MACE to optimize the Q, K inside of Dual Transformer Block. This modification ensures that our comparative analysis is conducted under a consistent and relevant framework. For EraseAnything, we adopt the official implementation settings.

\section{Additional Quantitative Results for Artistic Style Erasure}
Below are the results for erasing other two artistic styles. Our method similarly achieves the best $\mathrm{CLIP}_d$, which demonstrates superior decoupling capability for the target concepts.

\section{Additional Qualitative Results for Artistic Style Erasure}
\begin{figure*}[t] 
    \centering 
    \includegraphics[width=0.9\textwidth]{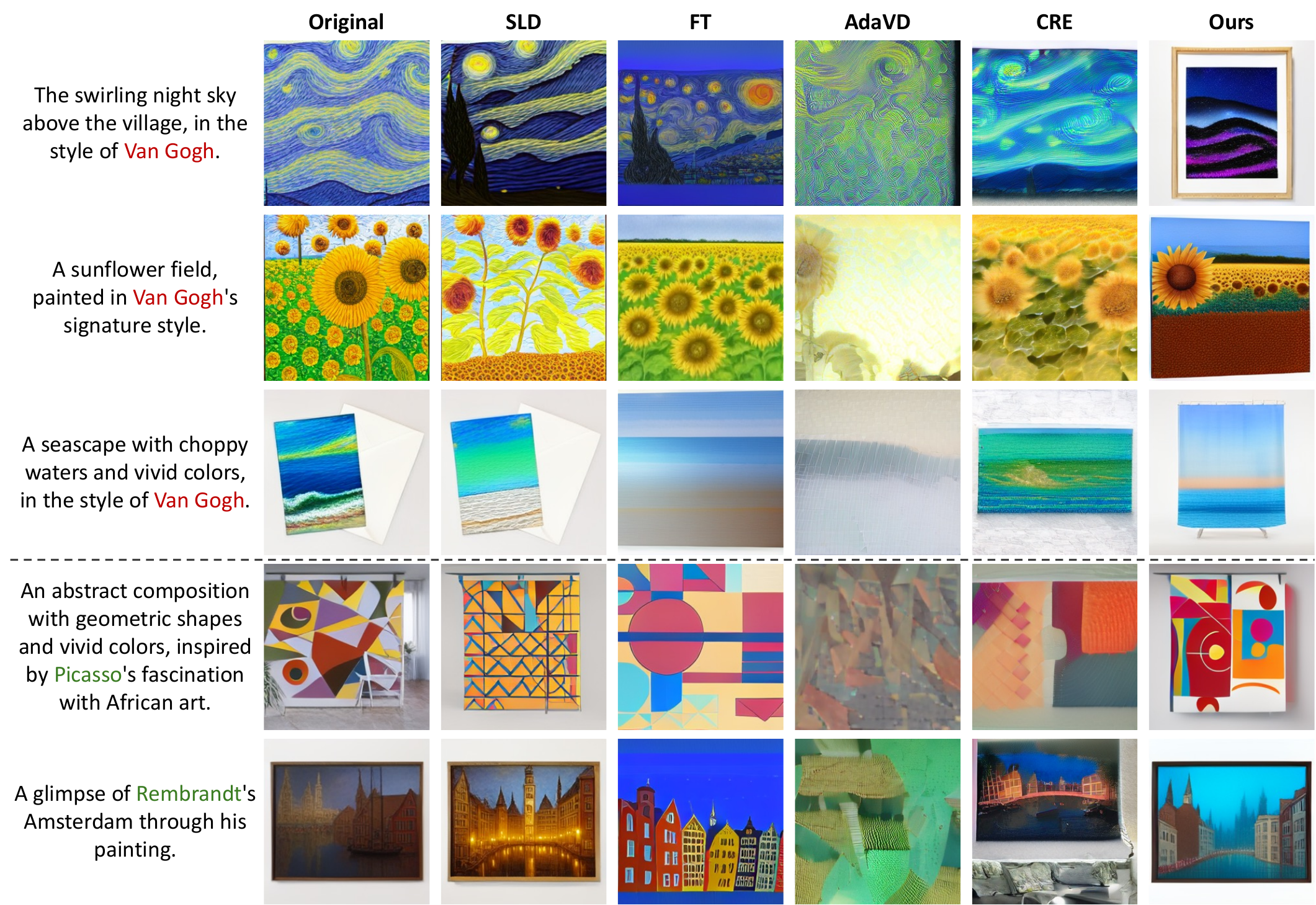} 
    \caption{Generated images after erasing ``Van Gogh'' style. Other artist styles such as ``Picasso'' and ``Rembrandt'' should be maintained.}
    \label{fig:van_gogh_erasure_appendix} 
\end{figure*}

Fig.~\ref{fig:van_gogh_erasure_appendix} presents qualitative results after erasing the ``Van Gogh'' style. Our method effectively removes the ``Van Gogh'' style while preserving non-target artistic styles.

\begin{table}[H]
\centering
\caption{CLIP scores for other artists. \textbf{Bold}: best. \underline{Underline}: second-best.}
\label{tab:metrics_artist_erasure_3}
\begin{adjustbox}{max width=\linewidth}
\begin{tabular}{lcccccc}
\toprule
\multirow{2}{*}{Method} & \multicolumn{3}{c}{Erasing \textit{``Rembrandt''}} & \multicolumn{3}{c}{Erasing \textit{``Caravaggio''}} \\ 
\cmidrule(lr){2-4} \cmidrule(lr){5-7} 
& $\mathrm{CLIP}_\mathrm{e}\downarrow$ & $\mathrm{CLIP}_\mathrm{u}\uparrow$ & $\mathrm{CLIP}_\mathrm{d}\uparrow$  
& $\mathrm{CLIP}_\mathrm{e}\downarrow$ & $\mathrm{CLIP}_\mathrm{u}\uparrow$ & $\mathrm{CLIP}_\mathrm{d}\uparrow$  
  \\ 
\midrule
SLD & 26.55 & \textbf{28.56} & 2.00 & 24.73& \textbf{28.18} & 3.45  \\
FT & \underline{21.89} & \underline{27.27} & \underline{5.40} & 22.09 & 26.24 & \underline{4.15}   \\ 
AdaVD & 22.36 & 21.91 & -0.45 & \textbf{19.80} & 22.52 & 2.72 \\
CRE & 24.18 & 23.65 & -0.53 & 22.57 & 24.33 & 1.76 \\
Ours & \textbf{20.25} & 26.85 & \textbf{6.60} & \underline{21.46} & \underline{26.45} & \textbf{4.99} \\
\midrule
LlamaGen & 25.77 & 28.34 & 2.57 & 25.23 & 28.22 & 2.99   \\
\bottomrule
\end{tabular}
\end{adjustbox}
\end{table}

\section{Prompts for the Refined Captioning Process}
Tab.~\ref{tab:prompt} and Tab.~\ref{tab:prompt2} are the system prompt and user prompt for executing Refined Captioning for Artistic Style Erasure. Tab.~\ref{tab:prompt3} and Tab.~\ref{tab:prompt4} are the system prompt and user prompt for executing Refined Captioning for Explicit Content Erasure. For the Object Removal task, as we observed that the generated images were relatively simple, containing only the corresponding objects and backgrounds, we directly used ``a picture'' to generate positive images without the Refined Captioning Process.

\begin{table*}[t]
    \centering
    \caption{Template of the system prompt for executing Refined Captioning for Artistic Style Erasure. \{artist\} will be replaced with the specific artist name. \{keywords\} will be replaced with the keywords for the artist.}
    \label{tab:prompt}
    \begin{tabular}{p{.96\linewidth}}
    \toprule
    You are an AI assistant specialized in conditional image captioning and prompt engineering. Follow these steps strictly: \\ 
    Step 1: Observe and analyze the provided image carefully. \\
    Step 2: Generate an initial caption describing only the realistic objects, scene, and actions depicted. Do not include artistic characteristics, styles, colors, or shapes related to {artist}. \\
    Step 3: Review the initial caption and remove any words or descriptions explicitly associated with \{artist\}'s style, such as \{keywords\}. \\
    Output format: \\
    Initial caption: "your initial caption here".Revised caption: "your revised caption here" \\
    \bottomrule
    \end{tabular}
\end{table*}

\begin{table*}[t]
    \centering
    \caption{Template of the user prompt for executing Refined Captioning for Artistic Style Erasure.}
    \label{tab:prompt2}
    \begin{tabular}{p{.96\linewidth}}
    \toprule
     Please follow the instructions carefully: First, annotate the image with an initial caption. Then, revise the caption to remove any style-specific terms related to the mentioned artist. Output clearly in the format provided by the system message. \\
    \bottomrule
    \end{tabular}
\end{table*}

\begin{table*}[t]
    \centering
    \caption{Template of the system prompt for executing Refined Captioning for Explicit Content Erasure.}
    \label{tab:prompt3}
    \begin{tabular}{p{.96\linewidth}}
    \toprule
     You are an AI assistant specialized in conditional image captioning. Follow these steps strictly: \\
     Step 1: Observe and analyze the provided image carefully. \\
     Step 2: Generate an initial caption describing the objects, scenes, and actions depicted, explicitly omitting any descriptions or references to nudity, sexually suggestive content, or exposed body parts. \\
     Step 3: Carefully review and ensure the caption does NOT contain any reference to nudity or sensitive areas. \\
     Output format: \\
     Initial caption: "your initial caption here"Revised caption: "your revised caption here"Important note: Never include any description or mention of nudity, sexually explicit or suggestive content, or exposed body parts." \\
    \bottomrule
    \end{tabular}
\end{table*}

\begin{table*}[t]
    \centering
    \caption{Template of the user prompt for executing Refined Captioning for Explicit Content Erasure.}
    \label{tab:prompt4}
    \begin{tabular}{p{.96\linewidth}}
    \toprule
     Please follow the instructions carefully: \\
     First, annotate the image with an initial caption. Then, revise the caption to completely remove any descriptions or mentions of nudity, sexually suggestive content, or exposed body parts. \\
     Output clearly in the format provided by the system message." \\
    \bottomrule
    \end{tabular}
\end{table*}

\end{document}